\documentclass{article}

\PassOptionsToPackage{numbers, compress}{natbib}

\usepackage[preprint]{neurips_2024}

\usepackage[utf8]{inputenc} 
\usepackage[T1]{fontenc}    
\usepackage{hyperref}       
\usepackage{url}            
\usepackage{booktabs}       
\usepackage{amsfonts}       
\usepackage{nicefrac}       
\usepackage{microtype}      
\usepackage{xcolor}         
\usepackage{graphicx}
\usepackage{amsmath}
\usepackage{adjustbox}
\usepackage{multirow}
\usepackage{wrapfig}
\usepackage{algorithm}
\usepackage{algpseudocode}
\usepackage{tabularx}
\usepackage{enumitem}
\usepackage{caption}

\usepackage{mathtools}

\usepackage{multirow}
\usepackage{makecell}

\usepackage{wrapfig}
\usepackage{floatflt}

\usepackage{tikz}
\usetikzlibrary{mindmap,trees,arrows.meta,positioning}

\title{Multimodal Federated Learning: A Survey through the Lens of Different FL Paradigms}

\author{%
  Yuanzhe Peng \\
  University of Florida \\
  Gainesville, FL 32611 \\
  \texttt{pengy1@ufl.edu} \\
  \And
  Jieming Bian  \\
  University of Florida \\
  Gainesville, FL 32611 \\
  \texttt{jieming.bian@ufl.edu} \\
  \And
  Lei Wang  \\
  University of Florida \\
  Gainesville, FL 32611 \\
  \texttt{leiwang1@ufl.edu} \\
  \And
  Yin Huang  \\
  University of Florida \\
  Gainesville, FL 32611 \\
  \texttt{yin.huang@ufl.edu} \\
  \And
  Jie Xu \\
  University of Florida \\
  Gainesville, FL 32611 \\
  \texttt{jie.xu@ufl.edu} \\
}

\begin{document}

\maketitle

\begin{abstract}
Multimodal Federated Learning (MFL) lies at the intersection of two pivotal research areas: leveraging complementary information from multiple modalities to improve downstream inference performance and enabling distributed training to enhance efficiency and preserve privacy. Despite the growing interest in MFL, there is currently no comprehensive taxonomy that organizes MFL through the lens of different Federated Learning (FL) paradigms. This perspective is important because multimodal data introduces distinct challenges across various FL settings. These challenges, including modality heterogeneity, privacy heterogeneity, and communication inefficiency, are fundamentally different from those encountered in traditional unimodal or non-FL scenarios.
In this paper, we systematically examine MFL within the context of three major FL paradigms: horizontal FL (HFL), vertical FL (VFL), and hybrid FL. For each paradigm, we present the problem formulation, review representative training algorithms, and highlight the most prominent challenge introduced by multimodal data in distributed settings. We also discuss open challenges and provide insights for future research. By establishing this taxonomy, we aim to uncover the novel challenges posed by multimodal data from the perspective of different FL paradigms and to offer a new lens through which to understand and advance the development of MFL.
\end{abstract}

\textbf{Keywords:} Multimodal Federated Learning, Horizontal Sample-space Partition, Vertical Feature-space Partition, Hybrid Partitioning, Modality Heterogeneity, Privacy and Security, Computational Efficiency, Communication Efficiency.

\section{Introduction}

\begin{wrapfigure}{r}{0.43\textwidth}
  \centering
  \includegraphics[width=0.75\textwidth, trim=350 100 150 70, clip]{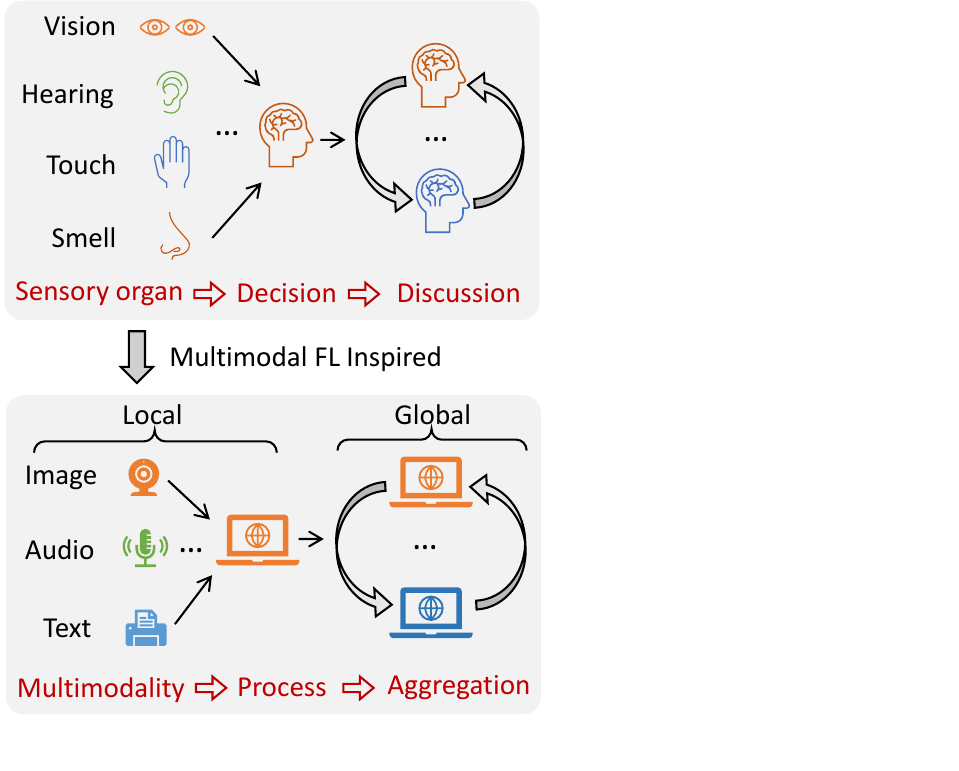}
  \vspace{-10mm}
  \caption{
MFL draws inspiration from human multi-sensory collaborative learning.
  }
  \label{intro}
\end{wrapfigure}

Humans perceive and understand the world through a rich blend of sensory inputs, including visual, auditory, tactile, and linguistic signals. This multimodal information is processed by the brain in a deeply integrated and context-aware manner, enabling comprehensive and nuanced decision-making, as shown in Fig. \ref{intro}. In a similar way, machine learning (ML) systems increasingly seek to process and integrate multimodal data to enhance prediction accuracy and robustness. With the proliferation of smart devices, wearable technology, and connected infrastructure, a vast volume of multimodal data is being generated every day. These data come from various sources such as cameras, microphones, inertial sensors, and text logs, often coexisting within a single application domain.

Multimodal learning (MML) has demonstrated impressive results in a wide range of applications, including image-text retrieval, video summarization, audio-visual speech recognition, emotion detection, and medical diagnosis. By aligning and fusing signals from multiple modalities, MML can capture richer semantics and improve the robustness of predictions in noisy or incomplete environments~\cite{wang2025cross,ghandi2023deep,hussain2021comprehensive}. However, in many practical scenarios, the collection and centralization of multimodal data are severely constrained by data silos, privacy regulations, and bandwidth limitations \cite{liang2021learning,dalmaz2022resvit}.
In social media platforms, for example, image, video, text, and audio data are often tightly coupled, yet they are generated on user-owned devices that contain highly sensitive personal content. In industrial environments, data from sensors deployed across different facilities may be owned by different business units or partners who are unwilling or unable to share raw data due to concerns over intellectual property or legal compliance. In the healthcare domain, clinical records, medical images, and physiological signals are distributed across hospitals and devices, often regulated by strict privacy policies such as the GDPR and HIPAA~\cite{yu2023multimodal,thrasher2023multimodal}. These trends call for learning frameworks that can collaboratively extract knowledge from distributed multimodal data without violating data sovereignty or privacy constraints.
Federated Learning (FL) addresses this need by enabling multiple clients to train a shared model collaboratively while keeping local data private~\cite{kairouz2021advances}. FL has demonstrated success in various unimodal tasks such as next-word prediction, image classification, and activity recognition~\cite{mcmahan2017communication,li2020federated,karimireddy2020scaffold,li2021fedbn,wang2020tackling}. Nevertheless, directly applying FL to multimodal learning is far from straightforward. The challenges encountered in Multimodal Federated Learning (MFL) are not simple extensions of those in unimodal FL or centralized MML, but rather represent new problem categories that emerge from the intersection of data heterogeneity, distributed computation, and modality-aware learning objectives.

\textbf{Emerging Challenges Unique to MFL.}  
Compared with unimodal FL, multimodal federated learning (MFL) involves a broader and more complex form of heterogeneity. Clients often hold different subsets of modalities due to differences in hardware capabilities, user preferences, or local data availability. This results in modality incompleteness and asynchronous modality participation across the network. If left unaddressed, such inconsistency can lead to biased model updates and reduced generalization performance. In addition, while unimodal FL typically assumes a consistent feature space across clients, MFL must accommodate variability in feature dimensionality, sampling frequency, and semantic interpretation. These factors make it more difficult to align representations and perform effective model aggregation.

In contrast to centralized multimodal learning, where modalities can be jointly processed and fused using globally shared representations, MFL lacks access to the full joint distribution of modalities. As a result, designing fusion strategies that capture cross-modal dependencies becomes more challenging. Since the relationships between modalities are distributed across clients, their integration must occur under limited coordination and incomplete information. Furthermore, the rich semantic content in multimodal data increases the risk of privacy leakage. In vertical MFL, for instance, intermediate feature embeddings from different modalities are transmitted to a central server for joint inference. This makes the system particularly vulnerable to inference attacks and reconstruction attempts that exploit correlations between modalities owned by different parties.

Hybrid FL settings introduce additional challenges, as both the sample space and feature space are partitioned. This configuration often involves modality-specific encoders that vary in computational demand, leading to inefficient training on resource-limited clients. Moreover, multi-branch models designed to handle different modality combinations typically require large communication overhead during synchronization. These issues collectively define a new class of challenges that are unique to MFL and cannot be addressed by simply extending existing unimodal or centralized approaches. A deeper, paradigm-aware understanding is essential for developing robust and efficient multimodal federated systems.

\begin{figure}
\centering
  \includegraphics[width=0.9\textwidth, trim=230 230 230 170,clip]{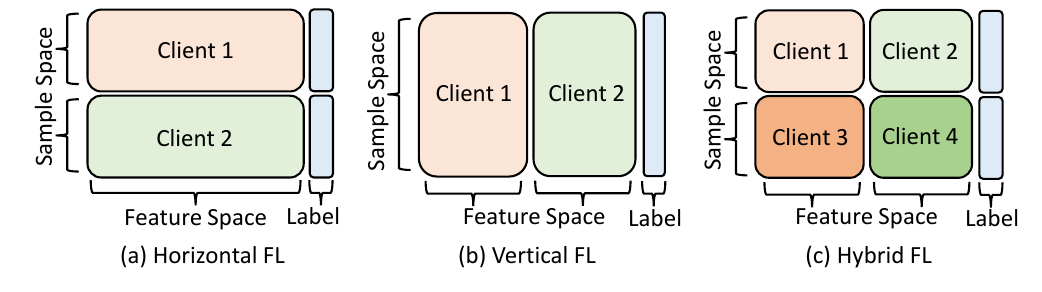}
  \caption{
(a) HFL addresses horizontally partitioned sample spaces with consistent feature spaces.  
(b) VFL addresses vertically partitioned feature spaces with consistent sample spaces.  
(c) Hybrid FL arises from partitioning both the sample space and the feature space. Note that all three paradigms discussed in this paper involve multimodal data, which introduces new challenges compared to traditional unimodal or non-FL settings.
  }
  \label{fig:diff}
\end{figure}

\textbf{Motivation.}  
Although several recent works have proposed algorithmic innovations in multimodal federated learning (MFL), there remains a lack of systematic organization that aligns these efforts with the foundational paradigms of federated learning (FL). In the conventional FL literature, three paradigms are widely recognized: horizontal FL (HFL), where clients share the same feature space but hold different data samples; vertical FL (VFL), where clients share the same samples but own different feature subsets; and hybrid FL, which involves partitioning both the feature and sample spaces, as illustrated in Fig.~\ref{fig:diff}. These paradigms impose different assumptions and constraints on data distribution, model architecture, and communication protocols.

Once multimodality is introduced, each of these FL paradigms gives rise to a distinct set of challenges. For example, HFL must address missing or incomplete modalities across clients, along with heterogeneous computational capabilities. VFL faces increased risks of privacy leakage due to the exchange of feature-level information across modalities. Hybrid FL presents both data fragmentation and heightened system complexity, as it requires simultaneous coordination of cross-client sample alignment and cross-modal feature fusion. Addressing these paradigm-specific challenges calls for a new taxonomy that organizes MFL research according to FL paradigms and explicitly links structural assumptions with the unique difficulties introduced by multimodal learning. One key question arises:

\textit{
How can we identify new challenges and uncover novel insights in MFL that do not arise in unimodal or non-federated settings, by exploring MFL under different FL paradigms?
}

To answer this question, we propose a paradigm-oriented taxonomy of MFL. This survey is the first to systematically organize MFL approaches based on the underlying FL paradigm, providing a structured framework that connects problem formulations, training strategies, and the core challenges introduced by multimodal data.

\textbf{Contributions.}  
Our key contributions are summarized as follows:

\begin{itemize}
    \item We present a novel taxonomy of existing MFL research grounded in three core FL paradigms: horizontal, vertical, and hybrid FL, which enables paradigm-aware comparisons and highlights how multimodality introduces new challenges.

    \item We provide problem formulations, representative algorithms, and a detailed analysis of the key challenges within each paradigm. We show that issues related to heterogeneity, privacy, and efficiency manifest differently across various FL settings.

    \item We compile a structured summary of publicly available datasets and real-world applications relevant to MFL, offering a practical reference for empirical validation and deployment.

    \item We identify open challenges and underexplored directions revealed through our taxonomy.
\end{itemize}

\textbf{Organization.}  
The remainder of this paper is organized as follows.  
Section~\ref{Proposed Taxonomy of MFL} introduces a new taxonomy of MFL based on three FL paradigms: horizontal, vertical, and hybrid FL. It motivates the need for paradigm-specific analysis and outlines the structure of each scenario.  
Section~\ref{Multimodal Horizontal Federated Learning} focuses on multimodal HFL. It formulates the problem, presents representative algorithms, and highlights modality heterogeneity as a key challenge.  
Section~\ref{Multimodal Vertical Federated Learning} covers multimodal VFL. It discusses the learning objectives and training strategies, with a focus on privacy risks arising from cross-party embedding transmission.  
Section~\ref{Multimodal Hybrid Federated Learning} explores multimodal hybrid FL, where both the feature and sample spaces are partitioned. This section emphasizes efficiency issues caused by the complexity of hybrid data distribution.  
Section~\ref{Applications and Datasets} introduces real-world applications and publicly available datasets relevant to MFL.  
Section~\ref{Open Challenges and Future Directions} outlines open challenges and future directions, including hardware heterogeneity, unsupervised learning, personalization, knowledge transfer, and interpretability.  
Section~\ref{Conclusion} concludes the paper with a summary of key insights and the significance of the proposed taxonomy.

\section{Proposed Taxonomy of MFL}
\label{Proposed Taxonomy of MFL}

To the best of our knowledge, no existing work has systematically classified MFL through the lens of different FL paradigms. Such a classification is essential because the presence of multimodal data introduces new challenges that are not encountered in unimodal FL or non-FL settings. In particular, the combination of modality heterogeneity, privacy concerns, and system efficiency constraints becomes much more complex when modalities are distributed across multiple clients.

Unlike unimodal FL, where data is typically homogeneous in structure and semantics, MFL involves multiple modalities such as image, audio, and text. These modalities differ in dimensionality, sampling frequency, and relevance to the learning task. When these differences are further fragmented across clients in a federated setting, the resulting optimization problem becomes significantly more difficult. Moreover, the nature of these challenges varies across different FL paradigms. 

In HFL (Fig.~\ref{fig:diff} a), each client owns a subset of data samples but shares a common feature space. Under multimodal settings, this means that clients are expected to process multiple modalities locally and collaboratively train shared models. A key challenge in this case is modality heterogeneity across clients. Some clients may have access to a full set of modalities, while others may only observe one or two. This creates imbalanced learning dynamics and can degrade the global model if not properly addressed. Moreover, different modalities often vary in their computational complexity. For example, clients processing video or spectrograms consume significantly more resources than those handling text or tabular data. As a result, ensuring fair participation and stable training under such heterogeneous modality availability and device capabilities becomes a central challenge.

In VFL (Fig.~\ref{fig:diff} b), clients hold different subsets of features for the same samples. When those features come from different modalities, the risk of privacy leakage increases significantly. Multimodal embeddings often carry rich semantic information that, when aligned across clients, may unintentionally reveal sensitive attributes even if each party only holds a partial view. This makes conventional encryption and aggregation techniques less effective for privacy protection. Furthermore, multimodal VFL requires tight coordination between clients to generate semantically consistent representations, which increases the potential for adversarial exploitation during intermediate representation exchange. Such vulnerabilities are particularly relevant in applications involving cross-institution collaboration, such as healthcare or finance, where each organization contributes a different modality but shares common user identifiers.

\begin{figure}
\centering
  \includegraphics[width=1\textwidth, trim=80 20 100 10,clip]{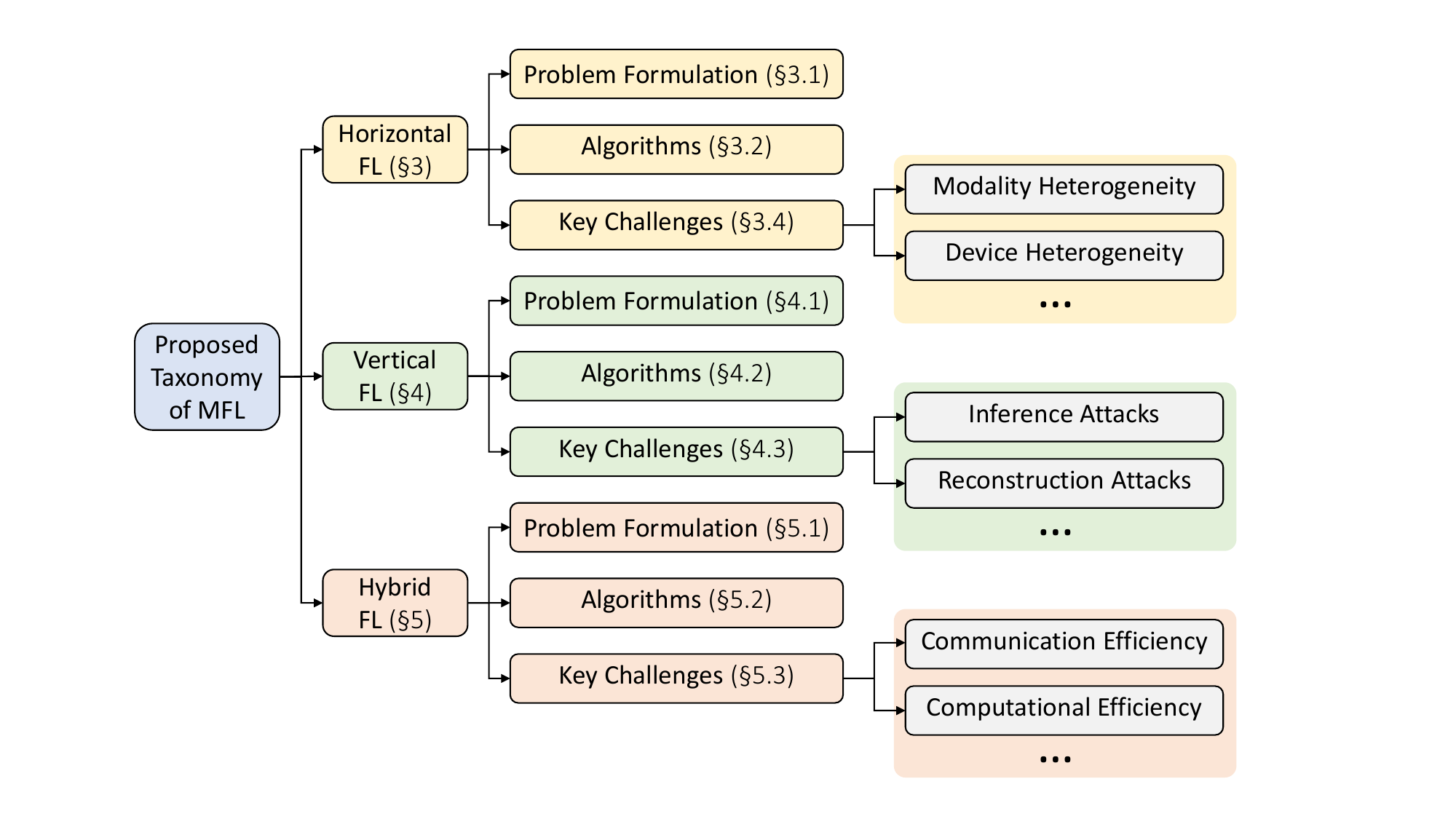}
  \caption{
Our proposed taxonomy presents MFL from the perspective of different FL paradigms. The key challenges we highlight are specific to the integration of multimodality within each FL paradigm, rather than general issues found in unimodal FL or centralized multimodal learning. 
  }
  \label{fig:taxonomy}
\end{figure}

In hybrid FL (Fig.~\ref{fig:diff} c), both the sample space and feature space are partitioned. This creates the most complex setting for MFL. Clients differ not only in the modalities they possess but also in the samples they observe. Coordinating training in such a setting requires addressing challenges related to both representation fusion and client synchronization. Efficiency becomes a primary concern, especially when modalities such as image and audio demand large memory and bandwidth resources. Communication overhead can quickly become a bottleneck, particularly when model architectures include modality-specific branches that must be updated separately. Moreover, hybrid FL systems often operate in heterogeneous environments where client devices vary significantly in connectivity and processing power. This further complicates the deployment of scalable and efficient MFL algorithms.

Given these observations, existing FL taxonomies that assume unimodal data are insufficient for describing the problem landscape faced by multimodal learning. To address this gap, we propose a comprehensive taxonomy of MFL structured around the three major FL paradigms. As shown in Fig.~\ref{fig:taxonomy}, each paradigm is analyzed along three dimensions: problem formulation, training algorithm, and key challenges. These challenges correspond to the key axes of heterogeneity, privacy, and efficiency, which we identify as the defining factors that govern the feasibility and performance of MFL systems. This taxonomy not only helps categorize prior work but also reveals new problem settings that are uniquely multimodal and federated in nature. By providing this structured understanding, our taxonomy lays the groundwork for future research that systematically addresses the trade-offs among modality diversity, user privacy, and system-level constraints in real-world MFL deployments.

\section{Multimodal Horizontal Federated Learning}
\label{Multimodal Horizontal Federated Learning}

We begin by introducing multimodal HFL, which serves as the foundation for many early studies in this domain. HFL assumes that clients share a common feature space but hold different data samples, making it a natural starting point for extending traditional unimodal FL to multimodal scenarios.

\subsection{Problem Formulation}  
\begin{wrapfigure}{r}{0.43\textwidth}
  \centering
  \includegraphics[width=0.9\textwidth, trim=360 200 180 140, clip]{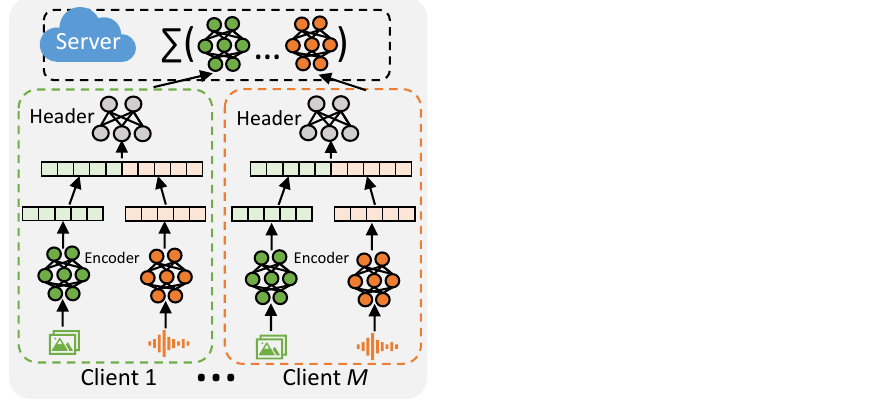}
  \caption{
Multimodal HFL, where each client shares the same multimodal feature space but holds a different sample space.
  }
  \label{fig_HFL}
\end{wrapfigure}

We consider a multimodal HFL system with \( M \) clients and one server, as shown in Fig.~\ref{fig_HFL}. Each client, indexed by \( m \) (\( m \in [M] \)), holds a local dataset \( \mathcal{D}_m = \{ (x_m^i, y_m^i) \}_{i=1}^{N_m} \), where \( x_m^i \) represents the multimodal input and \( y_m^i \) is the corresponding label for the \( i \)-th sample at client \( m \). All clients share the same set of modalities, and the dataset is horizontally partitioned, meaning that different clients hold different subsets of samples, but each sample contains the same types of features. The total number of samples across all clients is \( N = \sum_{m=1}^{M} N_m \).
Each client \( m \) maintains its own local model parameters \( \theta_m \), and its local objective is to minimize the empirical risk on its own dataset:  
$
f_m(\theta_m) := \frac{1}{N_m} \sum_{i=1}^{N_m} \ell(\theta_m; x_m^i, y_m^i),
$
where \( \ell(\theta_m; x_m^i, y_m^i) \) is the loss function that measures the prediction error of model \( \theta_m \) on sample \( (x_m^i, y_m^i) \).
The central server aims to optimize a global model \( \theta \) by aggregating local updates from the clients. The global objective function is defined as the weighted sum of local objectives across all clients:  
$
f(\theta) := \frac{1}{N} \sum_{m=1}^{M} N_m f_m(\theta).
$
\subsection{Training Algorithm}  
The HFL process involves the following steps:  
(1) \textit{Model Initialization}: The central server initializes the global model parameters \( \theta^{(0)} \) and broadcasts them to all clients.  
(2) \textit{Local Training}: Each client \( m \) receives the global model \( \theta^{(t)} \) at the beginning of round \( t \), and then performs local updates by minimizing its local objective \( f_m(\theta) \). Each client computes its local gradient:  
$
g_m^{(t)} := \nabla_{\theta} f_m(\theta^{(t)}),
$ 
and updates its local model using stochastic gradient descent:  
$
\theta_m^{(t+1)} := \theta^{(t)} - \eta g_m^{(t)},
$
where \( \eta \) is the learning rate.  
(3) \textit{Model Aggregation}: After local model updates, each client sends its updated model \( \theta_m^{(t+1)} \) to the central server. The server aggregates the models to update the global model as follows:  
$\theta^{(t+1)} := \frac{1}{N} \sum_{m=1}^{M} N_m \theta_m^{(t+1)}$.
This process is repeated for multiple communication rounds until convergence or a predefined stopping criterion is met.

\subsection{Representative Works on Multimodal HFL}

We now present representative works that adopt the HFL paradigm with multimodal data. Most of these studies essentially follow conventional multimodal learning principles in a straightforward manner, without fully addressing the unique challenges posed by the federated setting, such as statistical heterogeneity across clients, communication overhead, and the need for privacy-preserving modality alignment.

Agbley et al.~\cite{agbley2021multimodal} applied FL to melanoma detection using a dual-stream framework consisting of EfficientNet and a custom neural network to process medical images and clinical records separately. The outputs from the two branches are combined using a late-fusion strategy before parameter updates are shared with the server. The model achieved performance comparable to centralized training in terms of accuracy, F1, and AUC. However, this work primarily focused on handling modality heterogeneity and was evaluated on relatively simple downstream tasks.
Xiong et al.~\cite{xiong2022unified} proposed MMFed, a unified framework that combines co-attention mechanisms with personalization via Model-Agnostic Meta-Learning (MAML)~\cite{finn2017model}. Evaluated on video and signal datasets for human activity recognition~\cite{hu2023towards}, MMFed leverages co-attention to enhance modality complementarity and uses MAML to learn personalized models for each client. Although it demonstrates higher accuracy than traditional FL approaches, the model's performance is constrained by data imbalance, simplistic backbone networks, and increased communication costs introduced by the co-attention and meta-learning components.
Chen et al.~\cite{chen2024medical} developed a multimodal FL framework for automatic medical report generation. Their system employs CNN-based encoders (ResNet-101 and DenseNet-121) to extract visual features and incorporates a Transformer~\cite{vaswani2017attention} for textual generation by integrating image-derived features and predicted labels. Additionally, they proposed FedSW, a client scoring strategy that selectively updates model weights based on local performance. While the method outperforms local baselines in BLEU scores, it assumes homogeneous data quality across clients and imposes high computational demands due to its use of attention mechanisms.
Qi et al.~\cite{qi2023fl} introduced FL-FD, a framework designed for consistent-modality settings, which integrates one-dimensional time series data and camera-derived features. Time series inputs are converted into two-dimensional Gramian Angular Field representations, which are then stacked with visual data to form three-channel inputs. This approach enhances model performance and reduces complexity, making it well-suited for deployment in resource-constrained IoT systems. Nevertheless, its scalability to more diverse modality combinations and complex real-world tasks remains an open question.
Feng et al.~\cite{feng2023fedmultimodal} proposed FedMultimodal, a modular benchmark framework that decomposes the MFL pipeline into six key stages: data partitioning, feature processing, multimodal modeling, fusion strategies, FL optimization, and noise simulation. Designed for edge devices, the framework utilizes a lightweight Conv+RNN architecture and is compatible with standard FL algorithms, including FedAvg~\cite{mcmahan2017communication}, FedProx~\cite{li2020federated}, and FedOpt~\cite{reddi2020adaptive}. Despite its architectural flexibility, FedMultimodal achieves relatively low accuracy in image-text crisis detection tasks, likely due to limitations in the backbone’s feature extraction capabilities.

While the aforementioned works have explored multimodal extensions of HFL, most do not explicitly address the challenge of modality heterogeneity. Many assume that all clients have access to the same set of modalities, which does not hold in real-world scenarios where modality availability often varies due to device capabilities or privacy constraints. This oversight limits their applicability in practical settings. In our analysis, we find that modality heterogeneity is one of the most critical challenges introduced when extending HFL to multimodal data. We discuss this issue in greater depth in the following section.

\subsection{Key Challenge: Modality Heterogeneity}
A key challenge in multimodal HFL is modality heterogeneity, which refers to the diversity in modality types such as image, text, and audio, as well as their uneven distribution across clients. Each modality has distinct statistical properties and processing requirements. In decentralized settings, clients often possess only a subset of modalities, resulting in highly heterogeneous local feature spaces.
Unlike centralized multimodal learning, HFL must aggregate model updates from many clients with varying sensing capabilities, data availability, and modality combinations. Standard aggregation methods such as FedAvg assume a shared feature space across clients, which is often not valid in multimodal scenarios. Clients with non-overlapping modalities may learn incompatible representations, and directly aggregating their updates can introduce noise or conflicting gradients, which degrades global model performance and impedes convergence.

\begin{wrapfigure}{r}{0.5\textwidth}
    \centering
    \includegraphics[width=0.5\textwidth, trim=160 130 410 90, clip]{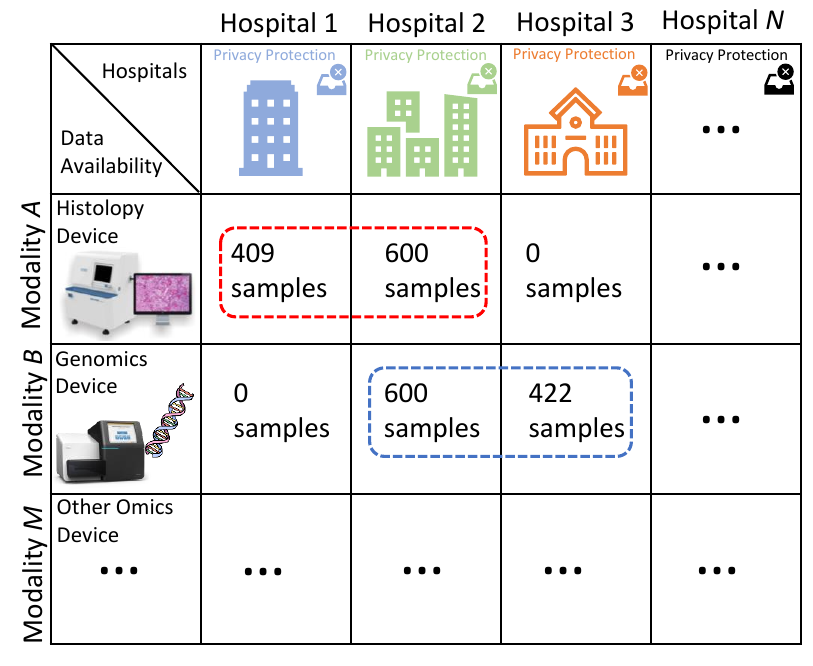}
    \caption{The problem of modality heterogeneity in computational pathology.}
    \label{fedmm}
\end{wrapfigure}

As shown in Fig.~\ref{fedmm}, computational pathology provides a representative example of this problem. 
Modality heterogeneity in this context arises from several factors. First, hardware limitations at different hospitals may restrict the use of certain diagnostic devices, leading to limited data modalities. Second, patient preferences for specific hospitals can result in uneven access to diagnostic data for the same disease. Third, data collected from different hospitals are typically non-identically and independently distributed, which further complicates training. These challenges make collaborative training difficult under standard FL frameworks. As a result, hospitals may be limited to local model training, missing the potential benefits of utilizing shared modalities across institutions. The problem becomes more pronounced as the number of clients increases, creating additional barriers to model alignment and scalability.
Addressing modality heterogeneity in HFL requires designing new aggregation strategies that account for modality differences, as well as developing learning architectures that adapt to heterogeneous input spaces.

Several other works address modality heterogeneity in HFL across different application domains. 
Yuan et al.~\cite{yuan2024communication} proposed mmFedMC, a modular decision-level fusion framework in which unimodal models are trained independently and combined during the decision-making stage. By selectively involving specific clients and modalities in each communication round, mmFedMC reduces communication overhead while mitigating semantic mismatches caused by modality inconsistency. However, the framework depends on numerous hyperparameters and demonstrates dataset-dependent performance, which limits its generalizability.
FedMSplit~\cite{chen2022fedmsplit} targets a more generalized scenario where both modality composition and task assignments are ambiguous. It partitions client data into modality-specific and globally shareable components using a graph-based attention mechanism, and guides aggregation through client similarity metrics. While this approach enables flexible handling of missing modalities, it introduces significant computational overhead on the server, particularly when aligning heterogeneous encoder outputs.
Chen and Zhang~\cite{chen2024disentanglement} introduced \textit{DisentAFL}, designed for modality-task agnostic federated learning. Their framework employs a two-stage process of knowledge disentanglement and semantic gating, enabling clients to exchange structured knowledge representations rather than raw gradients or embeddings. This helps align learning objectives across heterogeneous clients. While \textit{DisentAFL} consistently outperforms baselines, its complex architecture leads to increased training and communication overhead.
Zheng et al.~\cite{zheng2023autofed} proposed \textit{AutoFed}, which reconstructs missing modalities via inter-modal autoencoders. It uses cross-attention mechanisms for feature alignment and demonstrates strong performance in autonomous driving tasks. Nonetheless, AutoFed does not fully address concerns related to adversarial robustness, privacy guarantees, or incentive compatibility, which are critical in real-world deployment.
Ouyang et al.~\cite{ouyang2023harmony} proposed \textit{Harmony}, a two-stage federated training framework that first conducts modality-wise FL and then performs federated fusion learning to address modality-dependent training latency and enhance robustness to sensor failures. The resulting global model incorporates both unimodal and multimodal variants, achieving improved accuracy and efficiency over traditional baselines.

In sum, modality heterogeneity is the most critical challenge in multimodal HFL. It disrupts conventional assumptions regarding feature alignment, model compatibility, and communication efficiency. Addressing this challenge requires rethinking aggregation strategies, designing flexible representation alignment mechanisms, and incorporating personalized training pipelines.

\begin{table}
\centering
\caption{Summary of representative MFL methods.}
\resizebox{1\textwidth}{!}{  
\begin{tabular}{|c|l|l|l|}
\hline
\textbf{Category} & \textbf{Existing Work} & \textbf{Task} & \textbf{Key Idea} \\
\hline

\multirow{23}{*}{HFL} & MCARN \cite{yang2024cross} & Human activity recognition & Modality-collaborative network \\
 & FedFusion \cite{li2023fedfusion} & Remote sensing telemetry & Manifold learning and in-orbit fusion \\
 & FDARN \cite{yang2022cross} & Human activity recognition & A feature-disentangled network \\
 & FedMEMA \cite{dai2024federated} & Brain tumor segmentation & Modality-specific encoders and multimodal anchors \\
 & Melanoma \cite{agbley2021multimodal} & Melanoma detection & Late fusion by concatenating output features \\
 & AimNet \cite{liu2020federated} & Image-Text & Extract fine-grained representations \\
 & MMFed \cite{xiong2022unified} & Human activity recognition & Co-attention and personalization method \\
 & FedHGB \cite{chen2022towards} & Video classification & Hierarchical gradient blending \\
 & FedMultimodal \cite{feng2023fedmultimodal} & Benchmark & Decomposes training into six functional stages \\
 & FL-FD \cite{qi2023fl} & Human fall detection & Converts time-series signals into images \\
 & AutoFed \cite{zheng2023autofed} & Vehicle automation & Autoencoder-based data imputation \\
 & FedCLIP \cite{lu2023fedclip} & Image-Text & Adapter enhanced with attention mechanism \\
 & Harmony \cite{ouyang2023harmony} & Alzheimer’s monitoring & Modality-wise and federated fusion learning \\
 & FedCMR \cite{zong2021fedcmr} & Image-Text & Aggregates updates across shared subspaces \\
 & Mm-FedAvg \cite{zhao2022multimodal} & Human activity recognition & Autoencoder-based fusion \\
 & FedMEKT \cite{zhao2022multimodal} & Human activity recognition & Embedding knowledge transfer via distillation \\
 & CreamFL \cite{yu2023multimodal} & Image-Text & Contrastive learning and knowledge distillation \\
 & PFedPrompt \cite{guo2023pfedprompt} & Image-Text & Prompt tuning with personalized attention \\
 & PmcmFL \cite{bao2023multimodal} & Image-Text & Integrates a prototype memory mechanism \\
 & FedUSL \cite{yu2024fedusl} & Driving fatigue detection & Projects multimodal data into a unified latent space \\
 & FedMSplit \cite{chen2022fedmsplit} & Human activity recognition & Graph-based attention module \\
 & FedSea \cite{tan2023fedsea} & Search and classification & Domain adversarial alignment of features \\
 & mmFedMC \cite{yuan2024communication} & Multimodal healthcare & Selects optimal modalities and clients adaptively \\
 & DisentAFL \cite{chen2024disentanglement} & Multimodal generation & Disentangles asymmetric knowledge for symmetry \\
  & FedMM \cite{peng2024fedmm} & Computational pathology & HFL with separate unimodal feature extractors \\
   & Prio-modal \cite{bian2024prioritizing} & Human activity recognition & Adaptive modality scheduling for efficient training \\
\hline
\multirow{3}{*}{VFL} & MMVFL \cite{gong2023multi} & Image-Text & Two-step multimodal modeling for vertical FL \\
 & MVFL \cite{sundar2024toward} & Traffic analysis & Combines CCTV images and traffic tables \\
 & Fed-CRFD \cite{yan2024cross} & MRI reconstruction & Feature disentanglement with consistency regularization \\
\hline
\end{tabular}
}
\label{tab:hfl_summary}
\end{table}

\section{Multimodal Vertical Federated Learning}
\label{Multimodal Vertical Federated Learning}
In this section, we introduce VFL with multimodal data, a paradigm that differs fundamentally from HFL in terms of data partitioning and collaboration structure. In VFL, different parties hold complementary features (e.g., different modalities) corresponding to the same set of samples. This setup is common in cross-organization collaborations, such as between hospitals and insurance companies, or banks and FinTech platforms \cite{liu2024vertical}.

Current VFL methods that support multimodal data can be broadly categorized into two representative paradigms. The first includes recent works~\cite{castiglia2022compressed,castiglia2023flexible,castiglia2023less} that transmit embeddings and typically assume that both the server and clients have access to labels. The second, exemplified by earlier works such as~\cite{liu2022fedbcd}, relies on the transmission of partial gradients, designating one party (usually the server) as the active label holder, while the remaining parties act as passive participants without label access.
Our analysis in this section covers both categories of VFL frameworks. We discuss their respective problem formulations, training strategies, and key challenges.

\subsection{Problem Formulation}  
\begin{wrapfigure}{r}{0.3\textwidth}
  \centering
  \includegraphics[width=0.3\textwidth, trim=400 200 400 140, clip]{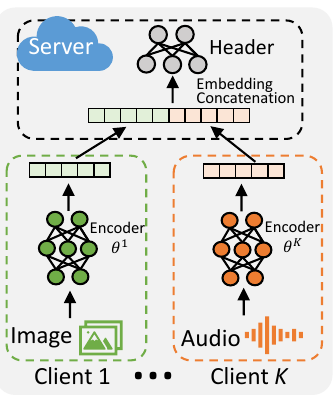}
  \caption{
Multimodal VFL. Each client trains its local model $\theta_k$ and uploads embeddings to the server, which concatenates the embeddings and feeds them into the server head model $\theta_0$.
  }
  \label{fig_VFL}
\end{wrapfigure}

We consider a multimodal VFL system consisting of \( K \) clients and a server, as shown in Fig.~\ref{fig_VFL}. The dataset \(\boldsymbol{x} \in \mathbb{R}^{N \times M}\) is vertically partitioned among the \( K \) clients across different feature spaces, where \( N \) denotes the number of data samples and \( M \) represents the number of features (e.g., multimodality).  
We define the local dataset for client \( k \) (\(k \in [K]\)) as \(\boldsymbol{x}_k \in \mathbb{R}^{N \times M_k}\), where \(M = \sum_{k=1}^{K} M_k\). The \( i \)-th row of \(\boldsymbol{x}\) corresponds to a data sample \( x^i \) (\( i \in [N] \)), where each sample \( x^i \) is composed of feature subsets held by each client, denoted as \( x_k^i \) for client \( k \), such that \( x^i = \{x_1^i, \dots, x_K^i\} \).  
Each sample \( x^i \) is associated with a global target task label \( y^i \), with each label corresponding to inherently distinct sensitive attributes across the \( K \) clients with different feature spaces.  
The server maintains a set of parameters \(\theta_0\) (referred to as the server head~\cite{collins2021exploiting}) and a loss function \(\ell(\cdot)\) for the global target task, which combines embeddings received from the \( K \) clients.  
Each client \(k\) locally holds a set of parameters \(\theta_k\) (referred to as the encoder) and an embedding function \(h_k(\cdot)\).  
The global objective is to minimize the following:
\begin{align}
\label{eqVFL}
\scalebox{1}{$
f(\Theta) :=  
\frac{1}{N} \sum_{i=1}^{N} \ell (\theta_0 \circ \{h_k (\theta_k; x_k^i)\}_{k=1}^K; y^i)
$}
\end{align}
where \(\Theta = \{\theta_0, \theta_1, \dots, \theta_K \}\) represents the global model.  
We set \(k=0\) as the server and define \(h_0(\theta_0; x^i) \coloneqq \theta_0\) for all \(x^i\) for simplicity. Let \(\nabla_k f(\Theta) := \frac{1}{N} \sum_{i=1}^N \nabla_{\theta_k} \ell ( \theta_0 \circ \{h_k(\theta_k; x_k^i)\}_{k=1}^K ; y^i)\) represent the partial derivatives of \(f (\Theta)\) with respect to the parameters \(\theta_k\) on client \(k\).  
Let \(\boldsymbol{x}^{\mathcal{B}}\) and \(\boldsymbol{y}^{\mathcal{B}}\) denote the samples and target task labels corresponding to a mini-batch \(\mathcal{B}\) of size \(B\). 
The stochastic partial derivatives for the parameters \(\theta_k\) are given by:
$\nabla_k f_{\mathcal{B}}(\Theta) :=  
\frac{1}{B} \sum_{x^i, y^i \in \boldsymbol{x}^{\mathcal{B}}, \boldsymbol{y}^{\mathcal{B}}} \nabla_{\theta_k} \ell (\theta_0 \circ \{h_k(\theta_k; x_k^i)\}_{k=1}^K ; y^i)$.

\subsection{Training Algorithm}  
\textbf{Idealized Training:} To illustrate the VFL training protocol, we first describe an idealized training algorithm, disregarding practical communication and computation constraints. 
The training proceeds in global rounds, where each round \( r \in [R] \) consists of local updates performed independently by each client.  
At the start of round \( r \), each client \( k \) initializes its local encoder as \( \theta_k^{r,0} = \theta_k^{r-1} \), using the converged local encoder from the previous round, and trains it on its entire local dataset \( \mathbf{x}_k \). 

The local training process involves applying (stochastic) gradient descent iteratively, updating \( \theta_k^{r,q} \) at each local step \( q \), and continuing until convergence.  
Since each client \( k \) requires embeddings, as well as the head model \( \theta_0 \), from other clients \( k' \neq k \) to compute its local partial gradients, information exchange is essential. However, in this idealized setup, communication occurs only at the end of each global round.  
Consequently, during round \( r \), client \( k \) has access only to the embeddings from other clients \( k' \), received in the previous round.  
Let \( h_k(\theta_k^{r}; \boldsymbol{x}_k) \) denote the embeddings generated by client \( k \) using its locally converged model \( \theta_k^r \) at the end of round \( r \). We define \( \Phi^r = \{ h_k(\theta_k^r; \boldsymbol{x}_k) \}_{k=1}^{K} \) as the collection of embeddings from all clients at round \( r \).  
During each local update step $q$ in round \( r \), client \( k \) updates its local encoder model based on: \textit{Fresh embeddings} \( h_k(\theta_k^{r,q}; \boldsymbol{x}_k) \), computed using its current encoder \( \theta_k^{r,q} \). \textit{Stale embeddings} from other clients, denoted as \( \Phi_{-k}^{r, 0} = \{ h_{k'}(\theta_{k'}^{r-1}; \boldsymbol{x}_{k'}) \}_{k' \neq k} \).  
At the end of each global round \( r \), all clients upload their embeddings to the server, which updates the server head model and redistributes the updated server head model along with the complete embeddings to the clients for the next training round.

\textbf{Practical Training:} The idealized training protocol, while conceptually clear, presents significant communication and computation challenges. Each global training round can be time-consuming, as encoders may require numerous local update steps to converge. Additionally, the communication overhead is substantial since embeddings for the entire dataset must be shared.  
To mitigate both communication and computation overhead, we introduce two practical modifications:  
(1) \textit{Mini-batch Training}: Instead of updating on the entire dataset, a mini-batch \(\mathcal{B}\) is sampled for training in each global round. To ensure timely embedding computation, sampling occurs at the end of the previous global round, allowing local clients to preprocess embeddings before the next round begins.  
(2) \textit{Fixed Local Update Steps}: Rather than waiting for local training to fully converge, we impose a fixed number of \( Q \) local updates per global training round. This constraint improves efficiency while maintaining effective model updates.
Moreover, although we assume that both the server and clients have copies of the labels, we also consider scenarios where these ideal conditions may not be met in practice.  
For instance, if labels are accessible only to one party (server), the label holder can still provide enough information for others to compute gradients for certain types of model architectures~\cite{liu2024vertical, castiglia2022compressed}.

\subsection{Key Challenge: Privacy Leakage}
One of the most critical challenges in multimodal VFL is privacy leakage. Unlike HFL, where clients hold the same features but different samples, VFL involves multiple parties that each possess different features of the same samples. These parties frequently exchange intermediate representations, such as embeddings, during training, which introduces new privacy vulnerabilities. Since no single party has access to the complete data, attackers can exploit observed information, such as partial intermediate representations, backpropagated gradients, or model predictions, to infer sensitive data held by distributed parties. Existing studies have identified four major categories of privacy attacks in VFL: representation-based inference, gradient-based reconstruction, label inference, and model stealing.

\textbf{Representation-based Inference Attacks:}  
This category includes attacks that exploit intermediate embeddings or model outputs to infer sensitive attributes or properties. Song et al.~\cite{song2019overlearning} demonstrated that models may inadvertently memorize sensitive attributes that are not part of the training labels. Property inference extends this idea by training classifiers on intermediate representations to recover group-level traits~\cite{melis2019exploiting, liu2022ml}. In VFL, the attack surface is typically limited to embeddings, which reduces but does not eliminate the risk of information leakage.
Luo et al.~\cite{luo2021feature} proposed several feature inference techniques tailored to different model types, including the Equality Solving Attack (ESA) for logistic regression, the Path Restriction Attack (PRA) for decision trees, and the Generative Regression Network (GRN) for complex models. Weng et al.~\cite{weng2020practical} and Hu et al.~\cite{hu2022vertical} introduced the Reverse Multiplication Attack (RMA) and the Protocol-aware Active Attack (PAA), both of which solve linear systems for feature recovery. Shadow modeling techniques were also developed by He et al.~\cite{he2019model} and Jiang et al.~\cite{jiang2022comprehensive}, where the attacker trains a surrogate model using auxiliary data.

\textbf{Gradient-based Reconstruction Attacks:}  
These attacks aim to reconstruct private input features or labels by exploiting gradients exchanged during training. In scenarios where attackers have access to sample-level gradients, He et al.~\cite{he2019model} and Jiang et al.~\cite{jiang2022comprehensive} proposed white-box model inversion (WMI) attacks in splitNN and aggregator-based VFL, respectively. CAFE~\cite{jin2021cafe} extended gradient inversion to the white-box setting and achieved high-quality recovery even with large batches. Gradient inversion methods, such as those proposed by Zou et al.~\cite{zou2022defending} and Kariyappa et al.~\cite{kariyappa2023exploit}, reconstruct inputs by minimizing the distance between estimated and real gradients. Tan et al.~\cite{tan2022residue} introduced Residue Reconstruction, solving gradient-matching problems to infer plaintext data from encrypted gradients. Although these methods have proven effective in horizontal FL, their applicability to VFL is often limited due to the unavailability of full gradients. For example, Yin et al.~\cite{yin2021see} proposed aligning estimated and observed gradients to recover training samples, but Xu et al.~\cite{xu2022cgir} showed that such approaches are infeasible when local models are not shared.

\begin{table}
\centering
\caption{Representative Attacks in Multimodal VFL}
\resizebox{1\textwidth}{!}{
\begin{tabular}{|l|l|l|}
\hline
\textbf{Category} & \textbf{Method} & \textbf{Key Idea} \\
\hline
\multirow{7}{*}{Representation-based Inference} 
  & ESA / PRA / GRN~\cite{luo2021feature} & Feature recovery methods for shallow and deep models \\
  & BFI~\cite{ye2024feature} & Infer binary features using leverage score sampling \\
  & RMA / PAA~\cite{weng2020practical, hu2022vertical} & Solve linear systems to recover features in logistic regression \\
  & RSA~\cite{weng2020practical} & Reveal feature ordering in tree-based models such as SecureBoost \\
  & Shadow Models~\cite{he2019model, jiang2022comprehensive} & Train surrogate model using auxiliary data \\
  & Attribute Inference~\cite{song2019overlearning} & Recover hidden attributes due to model overlearning \\
  & Property Inference~\cite{melis2019exploiting, liu2022ml} & Train classifiers over embeddings to infer group-level traits \\
\hline

\multirow{5}{*}{Gradient-based Reconstruction} 
  & WMI~\cite{he2019model, jiang2022comprehensive} & Optimize estimated inputs to match model outputs using gradients \\
  & CAFE~\cite{jin2021cafe} & White-box gradient inversion attack effective with large batches \\
  & GI~\cite{zou2022defending, kariyappa2023exploit} & Minimize distance between estimated and true gradients \\
  & RR~\cite{tan2022residue} & Solve gradient-matching problems to infer encrypted inputs \\
  & ~\cite{yin2021see, xu2022cgir} & Align gradients to reconstruct training data \\
\hline

\multirow{7}{*}{Label Inference} 
  & DLI~\cite{fu2022label} & Recover labels from sample-level gradient patterns \\
  & NS / DS~\cite{li2021label} & Score gradient norms or directions to infer binary labels \\
  & GI / RR~\cite{zou2022defending, tan2022residue} & Infer labels from batch-level gradients \\
  & PMC / AMC~\cite{fu2022label} & Fine-tune local models with auxiliary data or malicious optimization \\
  & SA~\cite{sun2022label} & Cluster model outputs to infer label groups without auxiliary data \\
  & LRI~\cite{qiu2022your} & Infer label relationships in graph tasks from prediction structures \\
\hline
\end{tabular}
}
\label{tab:privacy_attacks_vfl}
\end{table}

\textbf{Label Inference Attacks:}  
Labels held by the active party are also vulnerable. Fu et al.~\cite{fu2022label} proposed Direct Label Inference (DLI), which recovers labels from sample-level gradients. Li et al.~\cite{li2021label} showed that using a non-trainable output head, such as softmax, produces gradient patterns that clearly reveal label information. Even when the output layer is trainable, techniques such as Norm Scoring (NS) and Direction Scoring (DS) can still infer labels. When only batch-level gradients are available, Gradient Inversion~\cite{zou2022defending} and Residue Reconstruction~\cite{tan2022residue} remain effective.
In cases where gradients are not shared and only the final model is visible, model completion becomes the primary attack vector. Fu et al.~\cite{fu2022label} introduced Passive Model Completion (PMC), which uses auxiliary data to complete the model. Active variants of model completion manipulate the training process to maximize information leakage. Spectral Attack (SA)~\cite{sun2022label} bypasses the need for auxiliary labels by clustering model outputs. Label-related Relation Inference (LRI)~\cite{qiu2022your} targets relational patterns in graph-based tasks and infers label structures from model predictions.

\textbf{Model Stealing Attacks:}  
While most VFL attacks focus on compromising user data, model stealing aims to replicate the functionality of the target model. In centralized machine learning, adversaries may query the model to learn the input-output mapping~\cite{liu2022ml}. Although HFL shares the global model among participants, model stealing becomes more relevant in VFL, where no party has access to the full model. If a party observes consistent outputs across queries, it can approximate the model's decision boundary by training a surrogate model. This threat is particularly concerning in multimodal VFL, where partial exposure of multiple modalities may allow an attacker to capture complementary representations and reconstruct the logic of the target model.

Finally, we identify that VFL faces other challenges, such as computational efficiency, communication efficiency, and feature imbalance. However, these are common issues in distributed learning in general. Therefore, in this section, we focus on privacy leakage, which is particularly critical in VFL due to its inherent reliance on embedding transmission and feature space partitioning.

\section{Multimodal Hybrid Federated Learning}
\label{Multimodal Hybrid Federated Learning}
In this section, we introduce a new setting that cannot be effectively addressed by partitioning either the sample space or the feature space alone, which we refer to as hybrid FL. Although some studies use the term "hybrid FL" to describe scenarios where both the client and the server hold training data, in this paper, we define hybrid FL as the case where both the sample space and the feature space are partitioned across clients. When reduced to a single-dimensional partition, this setting corresponds to either the HFL or VFL cases discussed earlier.
Hybrid FL offers a more flexible paradigm for addressing real-world MFL challenges, as it accounts for both cross-client sample distribution and cross-party feature distribution.

\subsection{Problem Formulation}  
We consider a multimodal hybrid FL system with \( M \) silos, where each silo, indexed by \( m \) (\( m \in [M] \)), contains \( N_m \) samples. The total number of samples across all silos is \( N = \sum_{m=1}^{M} N_m \). Each silo represents a home or factory with a local hub and \( K \) distributed devices, each maintaining distinct feature spaces.  
These devices have sensors that collect various modalities, such as images and audio, which collectively have \( J \) sensors, where \( J \geq K \), capturing different features of the same sample. When \( K = J \), each device hosts one sensor.
For each silo \( m \), the local dataset \( \boldsymbol{x}_m \) is vertically partitioned across \( K \) AIoT devices along the feature space. The \( i \)-th row of \( \boldsymbol{x}_m \) denotes a data sample \( x_m^i \), defined as \( x_m^i := \{ x_m^{1, i}, \dots, x_m^{K, i} \} \), where each AIoT device \( k \) holds a disjoint yet complementary subset of features, denoted as \( x_m^{k, i} \).
Each sample \( x_m^i \) is associated with a label \( y_m^i \).  
Let \( \boldsymbol{y}_m \) denote all sample labels in silo \( m \), and \( \boldsymbol{x}_m^k \) represent the bi-orthogonally partitioned dataset maintained by AIoT device \( k \) in silo \( m \).
Each device \( k \) locally maintains a set of parameters \(\theta_m^k\) (referred to as the decomposed model) and an embedding function \( h_m^k(\cdot) \). The local hub (\( k=0 \)) maintains a set of parameters \(\theta_m^0\) (referred to as the head model \cite{collins2021exploiting}) and a loss function \(\ell(\cdot)\).  
Thus, the objective of silo \( m \) is to minimize:
\begin{align}
\scalebox{1}{$
f_m (\Theta_m ) :=   
\frac{1}{N_m} \sum_{i=1}^{N_m} \ell ( \theta_m^0 \circ \{ h_m^k (\theta_m^k; x_m^{k, i}) \}_{k=1}^{K} ; y_m^i ),
$}
\end{align}
where \(\Theta_m := \{ {\theta}_m^{0}, {\theta}_m^{1}, \dots, {\theta}_m^{K} \}\) represents the local composed model for silo \( m \). In silo \( m \), the partial gradient w.r.t. the coordinate partition \( \theta_m^k \) of device \( k \) can be expressed as:  
\begin{align}
\scalebox{1}{$
\nabla_k f_m (\Theta_m ) :=  
\frac{1}{N_m} \sum_{i=1}^{N_m} \nabla_{\theta_m^k} \ell ( \theta_m^0 \circ \{h_m^k (\theta_m^k; x_m^{k, i} ) \}_{k=1}^{K}; y_m^i).  
$}
\end{align}
The stochastic partial gradient of the coordinate partition \( \theta_m^k \) of device \( k \) can be expressed as:  
\begin{align}
\scalebox{1}{$
\nabla_k f_{m}(\Theta_m; \mathcal{B}_m) :=  
\frac{1}{B_m} \sum_{i \in \mathcal{B}_m} \nabla_{\theta_m^k} \ell ( \theta_m^0 \circ \{ h_m^k (\theta_m^k; x_m^{k, i} ) \}_{k=1}^{K}; y_m^i ), 
$}
\end{align}
where \( \mathcal{B}_m \) denotes a mini-batch of size \( B_m \).  
For brevity, we may omit \( \boldsymbol{x} \), \( \boldsymbol{y} \), \( \boldsymbol{x}_m \), and \( \boldsymbol{y}_m \) from \( f(\cdot) \) or \( f_{m}(\cdot) \).  
Additionally, we define  
$h_m^k(\theta_m^k; \boldsymbol{x}_m^{k, \mathcal{B}_m}) := \{h_m^k(\theta_m^k; x_m^{k, \mathcal{B}_m^1}), \dots, h_m^k(\theta_m^k; x_m^{k, \mathcal{B}_m^{B_m}})\}$  
as the set of embeddings from device \( k \) associated with the mini-batch \( \mathcal{B}_m \), where \( \mathcal{B}_m^i \) denotes the \( i \)-th sample in the mini-batch \( \mathcal{B}_m \).  
Moreover, we consider \( \nabla_k f_{\mathcal{B}_m}(\Theta_m) \) and  
$\nabla_k f_{\mathcal{B}_m} ( \theta_m^0,  h_m^1(\theta_m^1; \boldsymbol{x}_m^{1, \mathcal{B}_m}), \dots, h_m^K(\theta_m^K; \boldsymbol{x}_m^{K, \mathcal{B}_m}) )$  
to be equivalent and use them interchangeably.
Thus, the global objective is to minimize the following:  
\begin{align}  
\label{obj1}  
\scalebox{1}{$
f({\Theta}) := \frac{1}{N} \sum_{m=1}^M N_m f_m({\Theta}), 
$}
\end{align}  
where \( \Theta = \{ {\theta}^{0}, {\theta}^{1}, \ldots, {\theta}^{K} \} \) is the global composed model, and \( {\theta}^{k} \) denotes the decomposed model w.r.t. the split feature space \( k \).  
Objective~\eqref{obj1} evaluates how well the global composed model fits the multimodal system across \( K \) split feature spaces and \( M \) split sample spaces, which are bi-orthogonally partitioned.

\subsection{Training Algorithm}
The hybrid FL training algorithm is designed to enable flexible, scalable, and privacy-preserving learning across diverse multimodal environments, where both horizontal and vertical data partitioning coexist. 
It consists of four key components that are closely integrated to address the challenges posed by multimodal and non-i.i.d. data distributions.
(1) \textit{Flexible data partitioning} allows each device or silo to hold different types of information, such as disjoint feature subsets or sample partitions. Hybrid FL supports both horizontal and vertical configurations, depending on the local data structure and modality availability.
(2) \textit{Adaptive aggregation} is employed to integrate multimodal representations. The protocol supports multi-view or cross-modal fusion strategies, which aggregate heterogeneous embeddings generated by different parties into a unified representation.
(3) \textit{Mixed training strategies} allow for simultaneous horizontal and vertical FL participation. While some devices engage in horizontal FL, others collaborate through vertical FL. The global model is trained to integrate knowledge from both sources, facilitating generalization across sample and feature dimensions.
(4) \textit{Privacy-preserving mechanisms} are incorporated to ensure secure training and communication. Methods such as differential privacy \cite{wei2020federated, truex2020ldp}, secure aggregation \cite{fereidooni2021safelearn, so2022lightsecagg}, and homomorphic encryption \cite{zhang2020batchcrypt, zhang2022homomorphic} are applied to protect sensitive data while maintaining utility.

For an example of the hybrid FL training workflow introduced in~\cite{peng2024hybrid},  
each silo contains \( K \) vertical parties that engage in vertical FL, while \( M \) silos communicate horizontally through periodic global aggregation.  
Training begins with intra-silo vertical coordination. Every \( Q \) iterations, silo \( m \) samples a local mini-batch \( \mathcal{B}_m \) and initiates embedding exchange among its \( K \) parties.  
Each party \( k \) computes its local embedding \( h_m^k(\theta_m^{k, t}) \) based on its own modality and transmits it to the local hub.  
The hub aggregates these into a joint representation \( \Phi_m^{t_0} \) and broadcasts it to all parties within the silo.  
Each party then performs multiple local updates using stochastic partial gradients computed from \( \Phi_m^{k, t} \), which may include both current and cached components from earlier rounds.  

A key distinction between hybrid FL and traditional vertical FL methods such as FedBCD~\cite{liu2022fedbcd} lies in the inclusion of a trainable head model \( \theta_m^0 \) at each local hub.  
Inspired by the design principles in~\cite{collins2021exploiting}, this component supports advanced multimodal fusion architectures and enhances local decision-making.  

To address the non-i.i.d.\ nature of multimodal data across silos, hybrid FL incorporates global aggregation through horizontal FL.  
Specifically, every \( RQ \) updates, the global server aggregates the local models \( \theta_m^{k,t} \) across silos via:
$\theta^{k,t} = \frac{1}{N} \sum_{m=1}^M N_m \theta_m^{k,t}$,
where \( N_m \) is the number of local samples at silo \( m \).  
The updated global models are then distributed back to all silos.

In sum, unlike conventional MFL methods that incrementally treat multimodal inputs as a single high-dimensional feature vector, hybrid FL fundamentally decomposes training along both the feature and sample axes. This modular design enhances flexibility, promotes efficient learning, and better accommodates the structure of real-world multimodal data.

\subsection{Key Challenge: Efficiency}

\begin{wrapfigure}{r}{0.5\textwidth}
  \centering
  \includegraphics[width=0.5\textwidth, trim=310 165 310 155, clip]{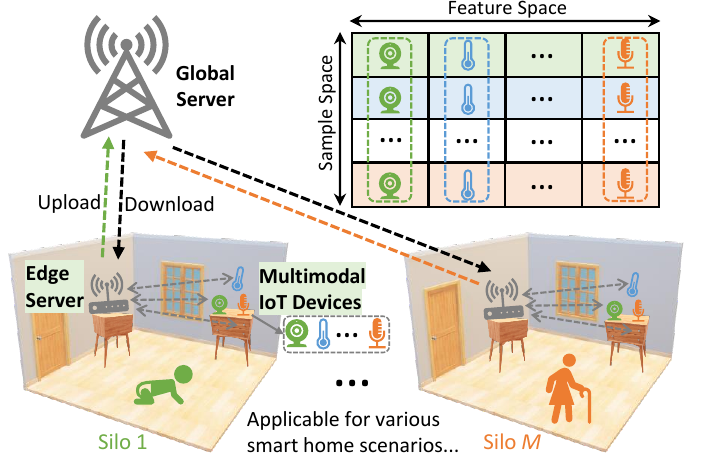}
  \caption{
    A multimodal system involves problem decomposition across feature and sample spaces.
  }
  \label{fig_multimodal IoT}
\end{wrapfigure}

While hybrid FL faces multiple challenges, including data heterogeneity and privacy leakage as discussed in earlier sections, efficiency stands out as the most fundamental and practical concern. This challenge primarily stems from the bi-orthogonal partitioning of data across both the sample and feature spaces, which introduces substantial system complexity in terms of computational cost and communication coordination.
The dual-axis training structure simultaneously involves VFL over partitioned features and HFL over distributed samples, as shown in Fig. \ref{fig_multimodal IoT}. 
This setup is further complicated by the presence of diverse data modalities, such as vision, language, audio, and time-series signals. These modalities vary in size, structure, and computational requirements, making it difficult to design a unified and efficient training pipeline. 

Unlike unimodal FL, where clients typically process similar data types, multimodal hybrid FL must accommodate heterogeneous devices holding complementary modalities.
Without effective resource management, hybrid FL systems may suffer from issues such as straggling devices, elevated energy consumption, and poor convergence. Coordinating communication and computation across both silos and feature holders is therefore essential to achieving scalability and training stability. 
We discuss the efficiency challenge below from both computational and communication perspectives.

\textbf{Computational Efficiency.}  
The computational cost in multimodal hybrid FL is substantial, as clients or devices must maintain decomposed local models, perform intermediate fusion, and participate in both local (intra-silo) and global (inter-silo) optimization. This is especially challenging for edge devices such as mobile phones, smart IoT units, and wearables, which typically have limited computational capacity and energy resources.

To address this issue, lightweight model design and adaptive training techniques are increasingly used. Model pruning~\cite{han2015deep} and quantization reduce memory usage and computation by removing redundant parameters or lowering the numerical precision of weights. Knowledge distillation~\cite{li2019fedmd, jeong2018communication} is also widely adopted to offload computationally intensive training tasks to more powerful servers, allowing smaller local models to benefit from shared knowledge across modalities and silos.
Split learning~\cite{vepakomma2018split} further extends the previously introduced VFL. In this approach, clients compute only the lower layers of the model and transmit intermediate representations to an edge server, which completes the remaining forward and backward propagation. This strategy shifts computation-intensive fusion and classification tasks from resource-constrained clients to nearby infrastructure, making it especially suitable for hybrid FL, where devices may contribute only specific modalities.
Recent approaches such as modality-aware scheduling \cite{bian2024prioritizing} and dynamic training have also shown promise. These techniques prioritize high-utility modalities and skip less informative ones, allowing computation to be allocated based on resource availability or modality relevance. This not only improves computational efficiency but also shortens convergence time and reduces energy consumption, both of which are essential in practical IoT applications \cite{peng2024hybrid, peng2024joint}.

\textbf{Communication Efficiency.}  
Hybrid FL imposes considerable communication overhead at two levels: intra-silo communication in VFL and inter-silo communication in HFL. Within each silo, vertical parties exchange modality-specific embeddings to support joint fusion and downstream inference. This becomes especially costly when dealing with high-dimensional data such as video frames or medical images. At the inter-silo level, the periodic aggregation of models for each modality introduces additional communication burden, particularly when silos maintain independent modality encoders and fusion modules.

To reduce communication costs, various compression techniques have been developed for federated learning. Sparse update methods such as Top-$k$ selection~\cite{aji2017sparse} and quantization strategies like QSGD~\cite{alistarh2017qsgd} transmit only the most informative gradients, significantly reducing transmission volume. Periodic update strategies, such as those used in FedPAQ~\cite{reisizadeh2020fedpaq}, allow clients to perform multiple local updates before synchronization. Variance-reduction techniques like SCAFFOLD~\cite{karimireddy2020scaffold} further stabilize training when communication is infrequent.
In hybrid FL, these methods can be adapted to modality-aware compression, where updates from large or less informative modalities are compressed more aggressively. Additionally, asynchronous FL protocols~\cite{xie2019asynchronous} and hierarchical aggregation schemes~\cite{liu2020client} effectively address heterogeneous update frequencies across vertical and horizontal participants. For example, HybridFed~\cite{tan2022towards} enables asynchronous aggregation of modality-specific sub-models, which improves convergence while supporting devices with varying capabilities and latency conditions.

\section{Applications and Datasets}
\label{Applications and Datasets}

In this section, we present representative real-world application scenarios and publicly available datasets that are applicable to MFL. It is important to note that most existing datasets originate from traditional centralized multimodal learning tasks. While these datasets are valuable for model benchmarking, new challenges arise when they are applied in distributed federated settings. In particular, issues such as sample-space partitioning (horizontal setting) and feature-space partitioning (vertical setting) must be carefully considered to reflect the heterogeneity and privacy constraints inherent in practical MFL deployments.

\subsection{Application Scenarios}

\textbf{Human Activity Recognition (HAR)} plays a pivotal role in ambient intelligence systems and has been widely adopted in applications ranging from health monitoring to smart homes and smart cities. HAR tasks typically rely on multimodal data sources such as RGB video, depth sensors, accelerometers, gyroscopes, audio signals, and radar sensors, offering a rich feature space for accurate activity classification. Traditional centralized learning approaches raise privacy concerns, particularly in sensitive environments such as healthcare or domestic monitoring. MFL provides an effective solution by enabling collaborative training across distributed devices or silos without requiring raw data sharing. Existing methods, including MMFed~\cite{xiong2022unified}, Mm-FedAvg~\cite{zhao2022multimodal}, FDARN~\cite{yang2022cross}, MCARN~\cite{yang2024cross}, and FedMEKT~\cite{le2025fedmekt}, have demonstrated promising results in preserving privacy while maintaining performance. A notable example is fall detection for elderly care~\cite{qi2023fl}, where local wearable devices and cameras collaboratively learn robust representations of fall patterns.

\textbf{Emotion Recognition} has broad applications in affective computing, driver monitoring, online education, and mental health assessment. Common multimodal signals include facial expressions, vocal tone, textual content, physiological indicators (such as heart rate and EEG), and behavioral cues (such as typing rhythm). MFL enables the personalization of emotion models using locally collected data, such as keystrokes and webcam feeds, without compromising user privacy. FedCMD~\cite{bano2024fedcmd} demonstrates a federated cross-modal distillation approach that uses unlabeled in-vehicle video to recognize driver emotions. In mental health contexts, combining keystroke dynamics with social media activity~\cite{liang2021learning} in MFL models could support early detection of depression or anxiety. Given the sensitive nature of such data, ethical safeguards such as informed consent and anonymization protocols must be applied~\cite{li2023deep}.

\textbf{Embodied Intelligence} focuses on equipping physical agents with cognitive capabilities for interacting with dynamic environments. These agents typically integrate multiple sensory modalities, including vision, audio, tactile feedback, and proprioception, to perceive and act effectively. Federated learning enables distributed training across a fleet of robots without transmitting raw sensory data, thereby preserving privacy. MFL extends this capability by supporting heterogeneous modality configurations across agents. For example, a robotic arm may utilize tactile and proprioceptive inputs, while a mobile robot may rely on visual and LiDAR data. Although research in this area is still emerging~\cite{gupta2021embodied}, MFL offers a scalable and privacy-preserving solution for cooperative robotic systems through parameter sharing and distributed training.

\textbf{Autonomous Driving} involves the generation and processing of vast amounts of multimodal data, including camera feeds, LiDAR scans, GPS signals, audio alerts, and driving behavior records. Public infrastructure further contributes modalities such as surveillance video and traffic signal information. MFL facilitates collaborative training across vehicles and edge nodes, enabling real-time adaptation and improved model generalization. AutoFed~\cite{zheng2023autofed} explores this concept by coordinating learning among vehicles without raw data exchange. MFL supports tasks such as object detection, scene understanding, and behavior prediction while maintaining privacy and reducing communication cost.

\textbf{Cross-Modal Retrieval} involves learning shared representations across modalities, such as matching images to textual descriptions. In federated settings, training such models requires alignment across heterogeneous data distributions. FedCMR~\cite{zong2021fedcmr} addresses client heterogeneity through weighted aggregation strategies that consider sample counts and label diversity. CreamFL~\cite{yu2023multimodal} leverages the Contrastive Language-Image Pretraining (CLIP) model~\cite{radford2021learning} and contrastive learning with shared public data to minimize privacy leakage. MFL frameworks enable collaboration among institutions or devices holding different modality types (for example, one with images and another with captions) to train powerful retrieval models while protecting raw data.

\textbf{Medical Diagnosis} is inherently multimodal, involving imaging modalities (such as CT, MRI, and ultrasound), structured EHRs, free-text clinical notes, genomics, and patient activity data. MFL frameworks allow for secure collaborative training across hospitals or departments. FedSW~\cite{chen2024medical} introduces a sliding window approach for sequential multimodal EHR data; FedMEMA~\cite{dai2024federated} incorporates attention-based modality fusion; and the melanoma diagnosis study~\cite{agbley2021multimodal} integrates clinical and dermoscopic images. MFL systems are capable of handling data heterogeneity across institutions, accommodating missing modalities, and enabling interpretable and robust diagnostic models.

\subsection{Publicly Available Datasets}
We introduce several publicly available datasets commonly used in different multimodal application scenarios. These datasets serve as valuable benchmarks for evaluating MFL frameworks under various data modalities and distribution settings.

\textbf{Multimodal Human Recognition (MHR)} refers to tasks that involve understanding human behavior through a combination of visual, auditory, textual, and sensor-based signals. These tasks include action recognition, activity detection, pose estimation, and behavior prediction~\cite{xiong2022unified, huang2020knowledge}. With the proliferation of smart devices, cameras, and wearables, MHR is becoming increasingly feasible in real-world settings. MFL enables the training of such models across distributed edge devices or user silos, ensuring privacy while leveraging diverse environments.

\textit{Kinetics-400}~\cite{kay2017kinetics} is a large-scale action recognition dataset containing more than 300,000 video clips, each approximately 10 seconds in length, spanning 400 human action categories. The videos are collected from YouTube and offer a wide range of real-world scenarios. In MFL experiments, clients can represent distributed camera networks or personal devices, supporting research on modality coverage, temporal segmentation, and label imbalance.

\textit{UCF101}~\cite{soomro2012ucf101} consists of 13,320 video clips classified into 101 categories of sports and daily activities. Most clips include only visual data, while some also contain audio, allowing research on robustness to missing modalities. In MFL settings, UCF101 can be partitioned by activity type, user ID, or data modality to simulate non-uniform sensor availability and partial modality learning.

\textit{UR Fall Detection Dataset}~\cite{kwolek2014human} is specifically designed for human fall recognition tasks. It includes 70 video sequences (30 fall events and 40 daily activities) captured using RGB and depth video from a Kinect sensor. This dataset is highly relevant to MFL applications in healthcare, particularly in eldercare monitoring systems where different clients (e.g., hospitals, homes, wearable manufacturers) collect distinct sensor streams and collaboratively train robust fall detection models using vertically partitioned data.

\textbf{Multimodal Emotion Recognition (MER)} focuses on identifying emotional states from diverse input modalities, including facial expressions, speech patterns, textual cues, and physiological signals. Emotions are closely tied to human decision-making and social interaction~\cite{zhao2021emotion}, making MER useful for social media analytics, psychological assessment, driver monitoring, and patient care. Federated approaches to MER allow models to be personalized using locally collected emotional data while preserving user privacy.

\begin{table}
\caption{Publicly available datasets across different application domains.}
\centering
\resizebox{1\textwidth}{!}{  
\begin{tabular}{|l|l|l|l|}
\hline
\textbf{Application} & \textbf{Dataset} & \textbf{Modality} & \textbf{Primary Task} \\
\hline
\multirow{5}{*}{Multimodal Human Recognition}
  & Kinetics-400 & Video, Text & Action Classification \\
  & UCF101 & Video, Audio, Text & Action Classification \\
  & UR Fall Detection Dataset & Image, Signal, Text & Fall Detection \\
  & WLASL & Video, Text & Sign Language Recognition \\
  & NTU RGB+D 120 & Video, Skeleton & Action Recognition \\
\hline
\multirow{5}{*}{Multimodal Emotion Recognition}
  & IEMOCAP & Video, Audio, Text & Emotion Classification \\
  & MELD & Video, Audio, Text & Emotion Classification \\
  & CMU-MOSEI & Video, Audio, Text & Emotion Classification \\
  & CMU-MOSI & Video, Audio, Text & Sentiment Analysis \\
  & SEWA & Video, Audio, Text & Affective State Recognition \\
\hline
\multirow{5}{*}{Vision-Language Models}
  & Flickr30k & Image, Text & Image Captioning, Retrieval \\
  & MS COCO & Image, Text & Object Detection, Captioning \\
  & VQA & Image, Text & Visual Question Answering \\
  & Visual Genome & Image, Text & Scene Graph Generation, QA \\
  & Conceptual Captions & Image, Text & Image Captioning \\
\hline
\multirow{5}{*}{Social Media Analysis}
  & Hateful Memes & Image, Text & Hate Speech Classification \\
  & UR-FUNNY & Video, Audio, Text & Humor Recognition \\
  & CrisisMMD & Image, Text & Crisis Event Classification \\
  & MMHS150K & Image, Text & Emotion and Sentiment Analysis \\
  & Memotion 2.0 & Image, Text & Humor and Sarcasm Detection \\
\hline
\multirow{2}{*}{Autonomous Vehicles}
  & Vehicle Sensor (custom) & Audio, Signal, Text & Driving Behavior Classification \\
  & nuScenes & Image, LiDAR, Radar, Text & Object Detection and Tracking \\
\hline
\multirow{3}{*}{Healthcare}
  & mHealth Dataset & Signal, Text & Activity Recognition \\
  & PTB-XL & Signal, Text & ECG Classification \\
  & MIMIC-CXR & Image, Text & Report Generation and Diagnosis \\
\hline
\multirow{2}{*}{Object Recognition}
  & ModelNet40 & Mesh, Point Cloud, Text & 3D Object Classification \\
  & ScanObjectNN & Point Cloud, Text & 3D Object Classification \\
\hline
\end{tabular}%
}
\label{tab:multimodal_datasets}
\end{table}

\begin{figure}
    \centering
    \includegraphics[width=1\textwidth, trim=280 15 300 20, clip]{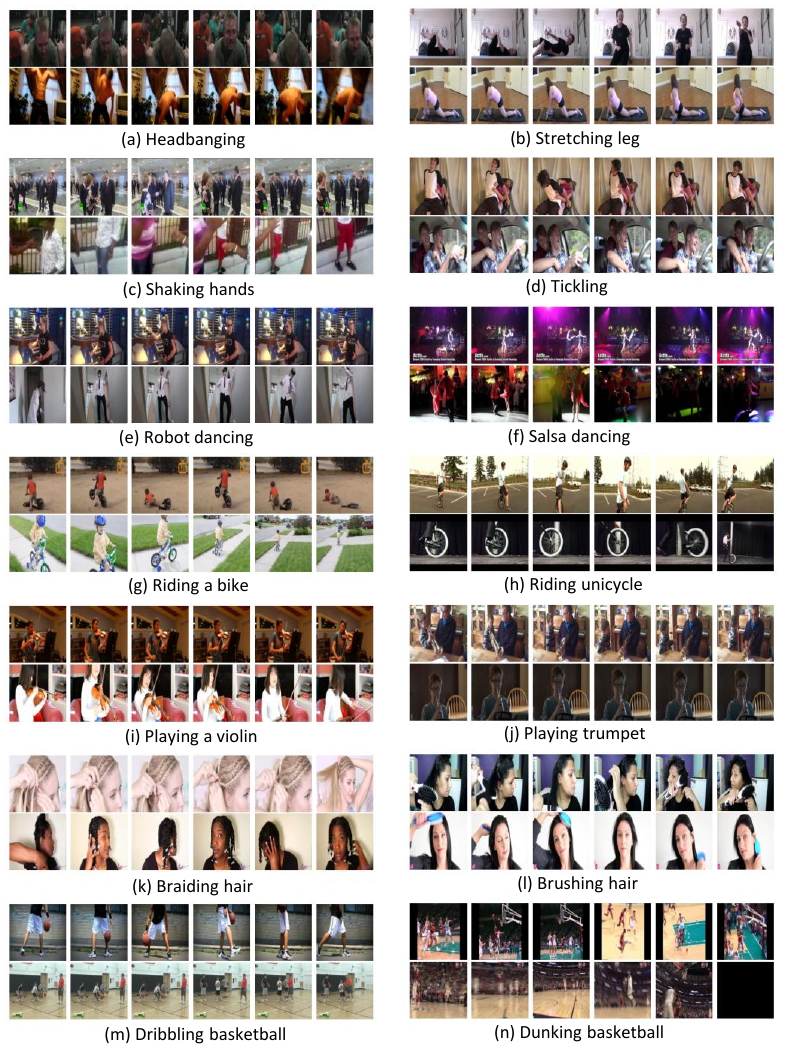}
    \caption{ Example classes from the Kinetics dataset \cite{kay2017kinetics}. 
Note that in some cases, a single image is not sufficient to recognize the action (e.g., “headbanging”) or to distinguish between classes (“dribbling basketball” vs. “dunking basketball”). The dataset includes: Singular Person Actions (e.g., “robot dancing”, “stretching leg”); Person-Person Actions (e.g., “shaking hands”, “tickling”); Person-Object Actions (e.g., “riding a bike”); same verb with different objects (e.g., “playing violin”, “playing trumpet”); and same object with different verbs (e.g., “dribbling basketball”, “dunking basketball”). These are realistic amateur videos, and there is often significant camera shake.
}
    \label{Kinetics}
\end{figure}

\textit{IEMOCAP}~\cite{busso2008iemocap} is a widely used dataset consisting of audiovisual and text recordings from ten professional actors engaged in scripted and improvised dialogues. It includes 151 sessions and 302 clips, totaling around 12 hours of content. Each segment is labeled with one of nine emotion categories, such as happiness, anger, and fear. 
\begin{wrapfigure}{r}{0.5\textwidth}
  \centering
  \includegraphics[width=0.5\textwidth, trim=170 100 190 90, clip]{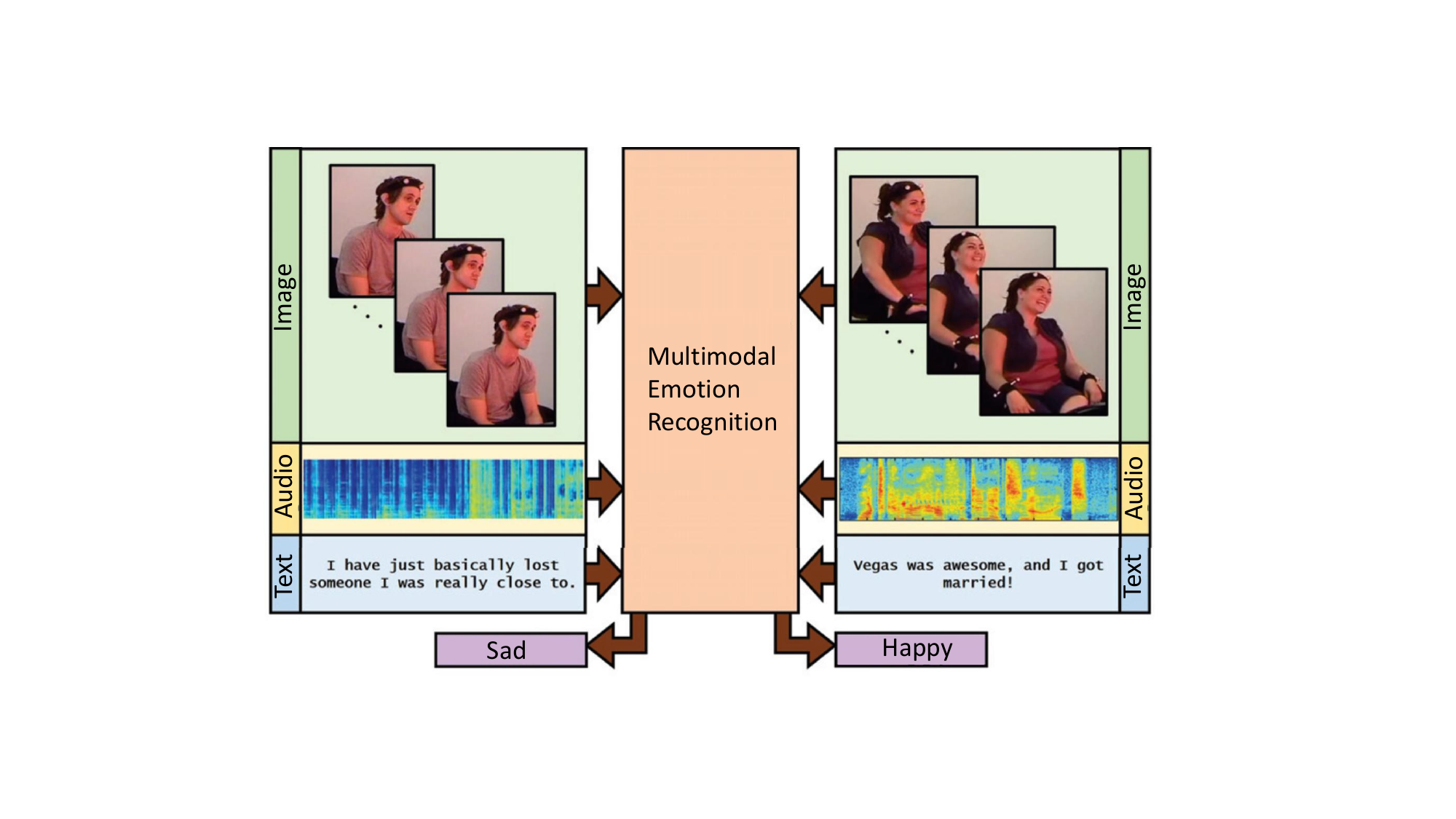}
  \caption{
Multimodal emotion recognition \cite{mittal2020m3er} based on IEMOCAP dataset.
  }
  \label{IEMOCAP}
\end{wrapfigure}
In MFL settings, clients can represent different participants or devices contributing partial modalities (e.g., only audio or only video), enabling research on cross-device and cross-view training under incomplete modality availability.

\textit{MELD}~\cite{poria2018meld} extends the EmotionLines dataset by incorporating visual and acoustic modalities. Derived from the TV show Friends, it contains over 13,000 utterances from 1,400 dialogues, annotated with seven emotion categories and sentiment polarity. MELD lends itself naturally to MFL, where each user may correspond to a unique speaker or dialogue participant, and local data distributions are highly non-IID.

\textit{CMU-MOSEI}~\cite{zadeh2018multimodal} contains more than 23,000 segmented monologue videos from over 1,000 speakers discussing 250 topics. Each segment is labeled with sentiment and six emotion tags. The large speaker diversity and topic coverage make it well-suited for federated partitioning, supporting research on personalization, speaker adaptation, and training under heterogeneous multimodal distributions.

\textbf{Vision-Language Models (VLMs)} have become a central topic in multimodal research due to advancements in joint image-text representation learning. With the rise of large-scale pretrained frameworks such as CLIP~\cite{radford2021learning}, ViLT~\cite{kim2021vilt}, and VLMo~\cite{bao2022vlmo}, integrating visual and textual modalities has become increasingly effective. These models support tasks such as cross-modal retrieval, caption generation, and visual reasoning. Several publicly available datasets serve as foundations for training and evaluating such models and also offer opportunities for MFL research by simulating non-IID modality distributions across clients.

\textit{Flickr30k}~\cite{plummer2015flickr30k} is a benchmark dataset consisting of over 31,000 images, each annotated with five natural language descriptions. 
The dataset covers a wide variety of scenes and cultural settings. 
\begin{wrapfigure}{r}{0.5\textwidth}
  \centering
  \includegraphics[width=0.5\textwidth, trim=1 1 1 1, clip]{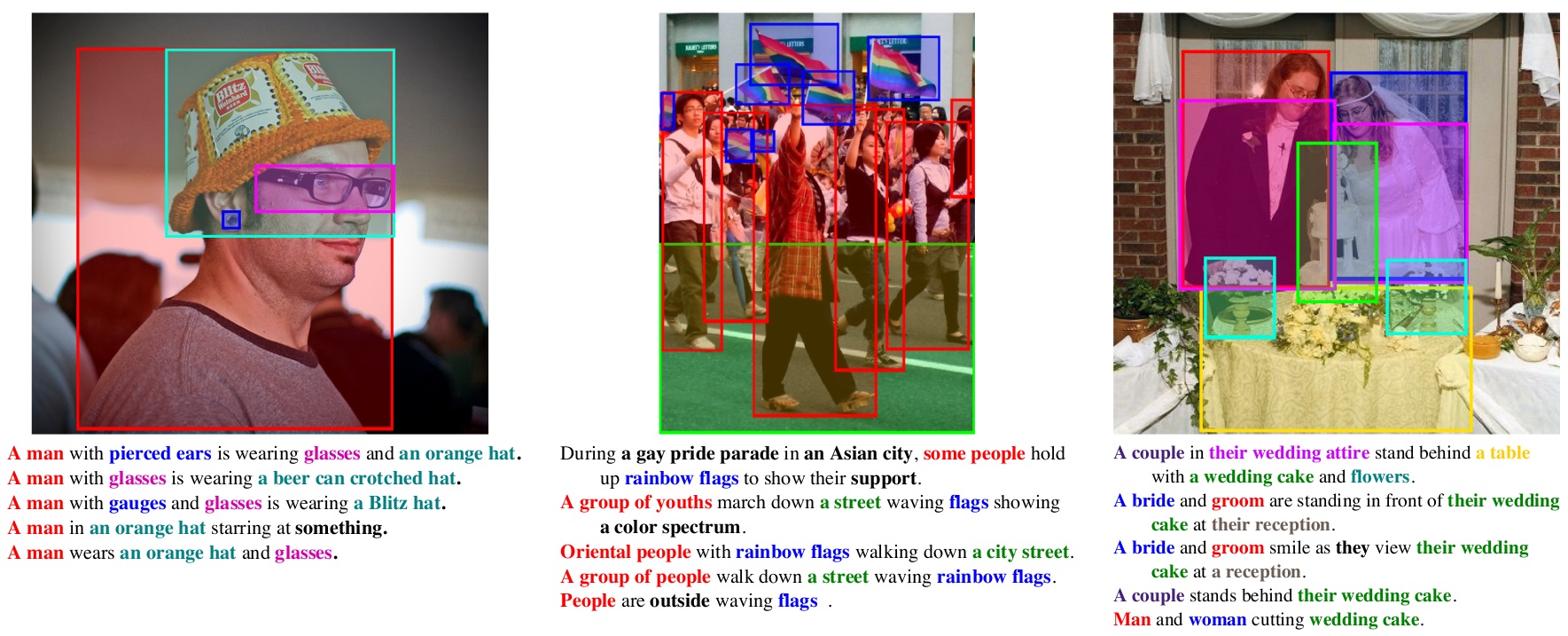}
  \caption{
Flickr30K dataset examples \cite{plummer2015flickr30k}.
  }
  \label{Flickr30K}
\end{wrapfigure}
In MFL scenarios, it can be used to simulate distributed content contributors, where each client holds a unique subset of image-caption pairs, enabling federated training of cross-modal alignment models.

\textit{MS COCO}~\cite{lin2014microsoft} is a large-scale dataset designed for various vision-language tasks, including object detection, segmentation, localization, and image captioning. It contains over 160,000 annotated images across training, validation, and test sets. In federated settings, clients can be assigned different subsets based on annotation types (e.g., some with captions, others with segmentation masks), supporting research on partial modality supervision and collaborative training across annotation silos.

\textit{VQA}~\cite{goyal2017making} (Visual Question Answering) is a well-known benchmark consisting of over 1.1 million image-question-answer triplets. Each image is paired with multiple questions designed to test visual reasoning capabilities. MFL applications may partition this dataset across clients based on question type or image domain, enabling personalized visual reasoning while preserving data privacy.

\section{Open Challenges and Future Directions}
\label{Open Challenges and Future Directions}

So far, we have summarized the primary challenges in MFL through the lens of different FL paradigms. It is important to clarify that our taxonomy highlights the novel challenges introduced by the integration of multimodality into federated settings, rather than the common issues typically encountered in conventional multimodal learning or unimodal FL scenarios.

Nevertheless, several open challenges remain that extend beyond the scope of our current taxonomy. These challenges can be viewed from alternative perspectives and reflect broader considerations within the field of MFL. In the following, we categorize and elaborate on key unresolved problems and outline promising directions for future investigation.

\textbf{System Heterogeneity and Resource Constraints}
MFL systems often suffer from significant device-level heterogeneity. Unlike traditional FL, where communication is the dominant bottleneck, MFL demands much more from local computation due to the complexity of multimodal data and models. Transformer-based architectures~\cite{vaswani2017attention}, frequently used in multimodal learning, are particularly resource-intensive and ill-suited for edge devices with limited computational power. Client selection methods based on resource availability~\cite{nishio2019client} may alleviate system load but raise concerns about fairness and inclusiveness. To address this, future research should focus on developing lightweight, modality-aware local models~\cite{jiang2023splitfed} and edge-assisted architectures that offload computations while maintaining data privacy. Balancing efficiency, fairness, and model performance in heterogeneous environments remains a core system-level challenge.

\textbf{Security, Privacy, and Trustworthiness}
MFL introduces amplified security risks due to both its federated structure and the information richness of multimodal data. Malicious clients can launch poisoning and backdoor attacks~\cite{bagdasaryan2020backdoor}, while adversaries may exploit multimodal correlations to perform inference or reconstruction attacks~\cite{fredrikson2015model, hitaj2017deep}. To enhance trustworthiness, future work should incorporate cryptographic techniques such as secure multi-party computation, homomorphic encryption~\cite{aono2017privacy}, and differential privacy~\cite{geyer2017differentially}. Blockchain-integrated frameworks~\cite{zhang2020blockchainfl} can also improve transparency and accountability. Additionally, moving away from centralized architectures and adopting peer-to-peer communication structures can help mitigate single points of failure and reduce trust assumptions.

\textbf{Label Scarcity and Unsupervised Learning}
Supervised learning remains the dominant paradigm in most MFL frameworks~\cite{chen2024disentanglement, chen2022fedmsplit, xiong2022unified}. However, real-world applications frequently involve partially or entirely unlabeled multimodal data, especially in scenarios such as sensor networks, healthcare, and ubiquitous computing. Semi-supervised MFL methods~\cite{zhao2022multimodal, yu2023multimodal, yu2024fedusl, le2025fedmekt} attempt to address this by employing pseudo-labeling, teacher-student mechanisms, or confidence-based filtering. A few pioneering works have explored unsupervised MFL setups~\cite{sun2022federated, saeed2020federated}, often through contrastive learning and transfer learning. Future work should design scalable, general-purpose unsupervised MFL frameworks inspired by centralized self-supervised strategies~\cite{chen2020simple, tian2020contrastive} and adapt semi-supervised FL schemes like FedMatch~\cite{jeong2020federated} to heterogeneous multimodal data.

\textbf{Personalized MFL}
Client-level variability in modality availability, data scale, and semantic distributions necessitates personalized solutions in MFL. In addition to post-hoc fine-tuning~\cite{arivazhagan2019federated} and multi-task optimization~\cite{smith2017federated}, several works now investigate explicit personalization mechanisms. DisentAFL and FedMSplit~\cite{chen2024disentanglement, chen2022fedmsplit} introduce shared-private modularity to extract general and client-specific representations. pFedPrompt~\cite{guo2023pfedprompt} proposes using client-level prompt vectors to guide learning over shared backbones, supporting personalized inference with minimal computation. The importance of personalized MFL continues to grow in domains like healthcare~\cite{tan2022towards}, recommendation systems~\cite{wang2023towards}, and smart devices, where user preferences and context vary significantly.

\textbf{Knowledge Transfer and Prompt-based Learning}
Knowledge transfer across modalities and clients is a compelling yet underexplored direction in MFL. Cross-modal distillation, where knowledge from one modality assists another, is especially beneficial when some modalities are missing or underrepresented. Existing work on unimodal distillation in MFL~\cite{zhang2022fedkd} can be extended into teacher-student multimodal pipelines. Additionally, recent developments in multimodal large language models (MLLMs) such as GPT-4o~\cite{openai2024gpt4o} enable prompt-based interaction across modalities. Prompt tuning offers communication-efficient and lightweight personalization. Future work should explore prompt learning in federated multimodal contexts, including federated prompt aggregation, visual-prompt alignment, and adapter sharing.

\textbf{Interpretability and Model Transparency}
Interpretability in MFL is essential for trust, safety, and model auditing, especially given the opacity of multimodal fusion processes. Unlike centralized systems where internal representations are readily inspected, federated setups limit visibility into both client data and model updates. Current techniques such as Grad-CAM~\cite{selvaraju2017grad} provide interpretability for image-text models but are insufficient for other modality combinations like sensor-audio or video-language. Future research should focus on cross-modal attribution methods, per-modality influence analysis, and interpretable client-side logging tools. Improving transparency will not only aid debugging and accountability but also enable secure deployment in sensitive domains such as healthcare and finance.

\section{Conclusion}  
\label{Conclusion}

In conclusion, MFL lies at the intersection of two central objectives in modern machine learning: leveraging complementary information across multiple modalities and enabling collaborative model training across distributed clients without exposing raw data. As the demand for privacy-preserving, cross-device intelligence continues to grow, MFL provides a promising framework for integrating multimodal data in distributed settings.
To better understand and address the unique challenges in this field, we propose a paradigm-oriented taxonomy that categorizes MFL methods into three foundational FL settings: HFL, VFL, and hybrid FL. Each paradigm introduces specific structural assumptions and data partitioning strategies that fundamentally influence how multimodal data are represented, processed, and shared. In multimodal HFL, clients often hold different subsets of modalities, which complicates feature alignment and global model consistency. In multimodal VFL, feature distribution across organizations raises privacy concerns during intermediate representation exchange. Hybrid FL, combining both sample and feature partitioning, further increases system complexity and coordination overhead.
Building on this taxonomy, we identify several open challenges that remain insufficiently addressed in the current literature. These include label scarcity, personalization under heterogeneous modality configurations, cross-modal knowledge transfer, and model interpretability. Effectively addressing these issues requires a nuanced understanding of how the underlying FL paradigm shapes the learning process.

By framing MFL through the lens of different FL paradigms, our work offers a structured perspective that not only clarifies the impact of data partitioning in multimodal settings but also highlights new challenges and research opportunities that are unique to MFL and rarely encountered in unimodal or centralized settings.


{
\bibliographystyle{IEEEtran}
\bibliography{ref}
}


\end{document}